\documentclass[11pt,a4paper]{article}
\usepackage[hyperref]{emnlp2020}
\usepackage{times}
\usepackage{multirow}
\usepackage{latexsym}
\usepackage{stfloats}
\usepackage[ruled,vlined]{algorithm2e}
\usepackage{tikz}

\usepackage{microtype}

\makeatletter
\newcommand{\removelatexerror}{\let\@latex@error\@gobble}
\makeatother

\title{Focus on what matters: \\ Applying Discourse Coherence Theory to Cross Document Coreference}

\author{William Held\textsuperscript{1}, Dan Iter\textsuperscript{2}, Dan Jurafsky\textsuperscript{2} \\
  \textsuperscript{1}Sunshine Products \\
  \textsuperscript{2}Department of Computer Science, Stanford University\\
  \texttt{will@sunshine.com, \{daniter,jurafsky\}@stanford.edu} \\
  }

\date{}

\begin{document}
\maketitle
\begin{abstract}
       Performing event and entity coreference resolution across documents vastly increases the number of candidate mentions, making it intractable to do the full $n^2$ pairwise comparisons. Existing approaches simplify by considering coreference only within document clusters, but this fails to handle inter-cluster coreference, common in many applications. As a result cross-document coreference algorithms are rarely applied to downstream tasks. We draw on an insight from discourse coherence theory: potential coreferences are constrained by the reader's discourse focus. We model the entities/events in a reader's focus as a neighborhood within a learned latent embedding space which minimizes the distance between mentions and the centroids of their gold coreference clusters. We then use these neighborhoods to sample only hard negatives to train a fine-grained classifier on mention pairs and their local discourse features. Our approach\footnote{Code is available at \url{https://github.com/Helw150/event_entity_coref_ecb_plus}} achieves state-of-the-art results for both events and entities on the ECB+, Gun Violence, Football Coreference, and Cross-Domain Cross-Document Coreference corpora. Furthermore, training on multiple corpora improves average performance across all datasets by 17.2 F1 points, leading to a robust coreference resolution model for use in downstream tasks where link distribution is unknown.
\end{abstract}

\begin{figure*}[htp]
    \centering
    \includegraphics[width=15cm]{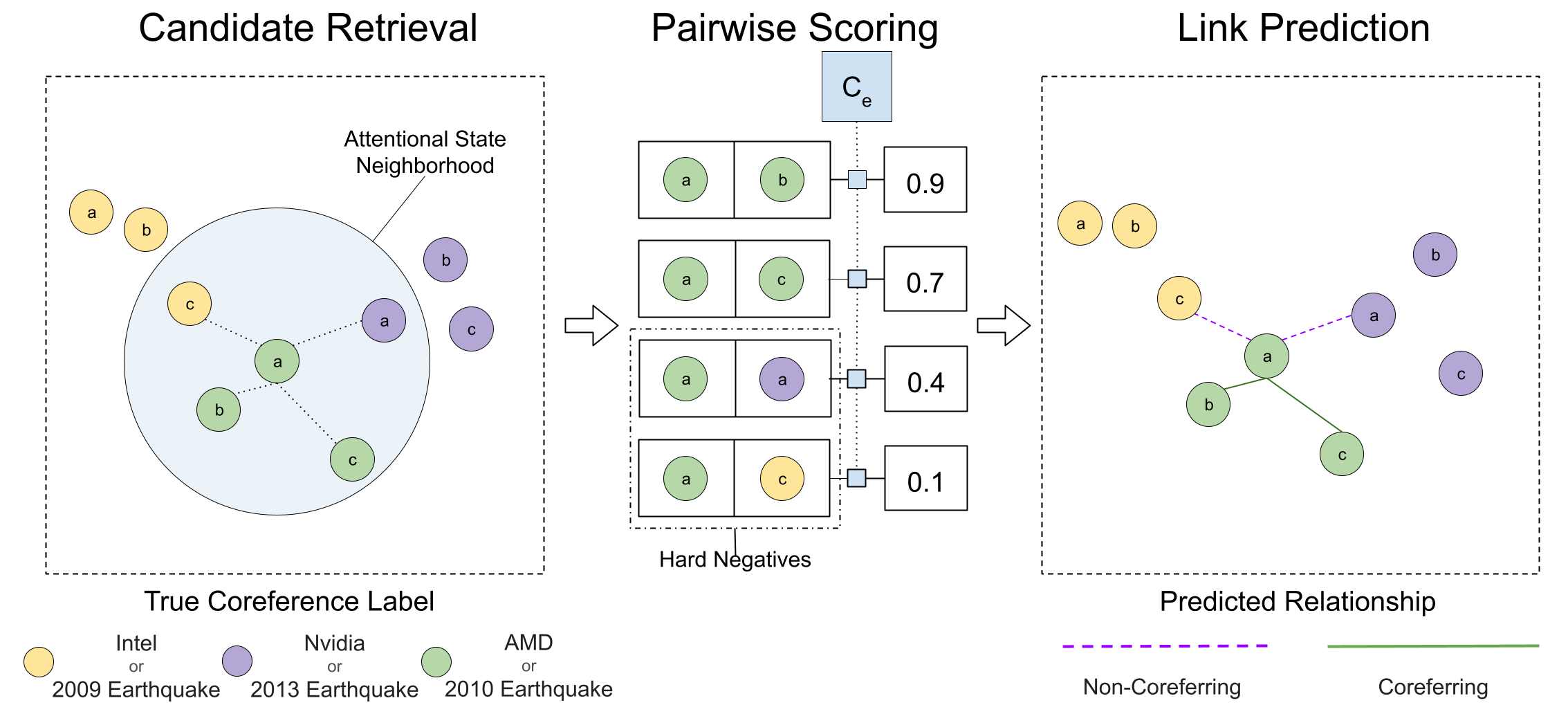}
    \caption{A high level overview of our system: For a particular mention, candidate coreferring mentions are retrieved from a neighborhood surrounding the mention. These candidate pairs are fed to a pairwise classifier specialized for hard negatives fetched from this space. This allows our method to create a high fidelity coreference graph with minimal pairwise comparison and no a priori assumptions about coreference. We use a bi-encoder for candidate retrieval and a cross-encoder for pairwise classification~\citep{humeau2020polyencoders}.}
    \label{overview}
\end{figure*}

\section{Introduction}
Cross-document coreference resolution of entities and events (CDCR) is an increasingly important problem, as downstream tasks that benefit from coreference annotations --- such as question answering, information extraction, and summarization --- begin interpreting multiple documents simultaneously. Yet the number of candidate mentions across documents makes  evaluating the full $n^2$ pairwise comparisons intractable~\citep{cremisini-finlayson-2020-new}. For single-document coreference,  the search space is pruned with simple recency-based heuristics, but there is no natural corollary to recency with multiple documents. 

Most CDCR systems  thus instead \textit{cluster} the documents and perform the full $n^2$ comparisons only within each cluster, disregarding inter-cluster coreference~\citep{lee-etal-2012-joint, yang2015hierarchical, choubey-huang-2017-event, barhom2019revisiting, cattan2020streamlining, yu2020paired, caciularu2021crossdocument}. This was effective for the ECB+ dataset, on which most CDCR methods have been evaluated, because ECB+ has lexically distinct topics with almost no inter-cluster coreference.
 
Such document clustering, however, keeps CDCR systems from being generally applicable. \citet{Bugert2020BreakingTS} shows that inter-cluster coreference makes up the majority of coreference in many applications. \citet{cremisini-finlayson-2020-new} note that 
document clustering methods are also unlikely to generalize well to real data where documents lack the significant lexical differences of ECB+ topics. These issues present a major barrier for the general applicability of CDCR.

Human readers, by contrast, are able to perform coreference resolution with minimal pairwise comparisons. How do they do it? Discourse coherence theory~\citep{Grosz1977TheRA, grosz-1978-focusing-dialog, grosz-sidner-1986-attention} proposes a simple mechanism: a reader \textit{focuses} 
on only a small set of entities/events from their full knowledge. This set,  the \textit{attentional state}, is constructed as entities/events are brought into focus either explicitly by reference or implicitly by their similarity to what has been referenced. Since attentional state is inherently dynamic --- entities/events  come into and out of focus as discourse progresses --- a document level approach is a poor model of this mechanism.

We propose modeling focus at the mention level using the two stage approach illustrated in Figure \ref{overview}. We model attentional state as the set of K nearest neighbors within a latent embedding space for mentions. This space is learned with a distance based classification loss to construct embeddings that minimize the distance between mentions and the centroid of all mentions which share their reference class.

These attentional state neighborhoods aggressively constrain the search space for our second stage pairwise classifier. This classifier utilizes cross-attention between mention pairs and  their  local  discourse  features to capture the features important within an attentional state which are comparison specific~\citep{grosz-1978-focusing-dialog}. By sampling from attentional state neighborhoods at training time, we train on only hard negatives such as shown in Table \ref{fig:examples}. We analyze the contribution of the local discourse features to our approach, providing an explanation for the empirical effectiveness of our classifier and that of earlier work like \citet{caciularu2021crossdocument}.

Following the recommendations of \citet{bugert2020crossdocument}, we evaluate our method on multiple event and entity CDCR corpora, as well as on cross-corpus transfer for event CDCR. Our method achieves state-of-the-art results on the ECB+ corpus for both events (+0.2 F1) and entities (+0.7 F1), the Gun Violence Corpus (+11.3 F1), the Football Coreference Corpus (+13.3 F1), and the Cross-Domain Cross-Document Coreference Corpus (+34.5 F1). We further improve average results by training across all event CDCR corpora, leading to a 17.2 F1 improvement for average performance across all tasks. Our robust model makes it feasible to apply CDCR to a wide variety  of downstream tasks, without requiring  expensive new coreference annotations to enable fine-tuning on each new corpus.
(This  has been a huge effort for the few tasks that have attempted it like  multi-hop QA~\citep{dhingra-etal-2018-neural, DBLP:journals/corr/abs-1910-02610} and multi-document summarization~\citep{falke-etal-2017-concept}.)

\begin{table*}[]
\centering
\small
\begin{tabular}{|c|l|c|}
\hline
Mention Type       & \multicolumn{1}{c|}{Mention}                                                                                & Relationship    \\ \hline
\multirow{3}{*}{Event}  & A preliminary magnitude of 2.0 \textbf{struck} near The Geysers                                             & Root            \\ \cline{2-3} 
                        & The earthquake \textbf{struck} at about 7:30 a.m                                                            & Coreferring     \\ \cline{2-3} 
                        & The temblor \textbf{occurred} at 9:27 a.m                                                                   & Different \\ \hline
\multirow{3}{*}{Entity} & ... would turn \textbf{AMD} into one of the world's largest providers of graphics chips.                                       & Root            \\ \cline{2-3} 
                        & ... \textbf{the company} announced that they have reached a \$334 million agreement  & Coreferring     \\ \cline{2-3} 
                        & Intel, the world's largest \textbf{graphics-chipmaker}, declined to comment... on the deal.                             & Different \\ \hline
\end{tabular}
\caption{Examples of positives and hard negatives within an attentional state neighborhood.}
\label{fig:examples}
\end{table*}
\begin{table*}[]
\centering
\small
\setlength{\tabcolsep}{4pt}
\begin{tabular}{l|ccc|ccc|ccc|ccc|ccc|}
\cline{2-16}
                                                         & \multicolumn{9}{c|}{Events}                                                                                                                   & \multicolumn{6}{c|}{Entities}                                                                 \\ \cline{2-16} 
                                                         & \multicolumn{3}{c|}{ECB+}                     & \multicolumn{3}{c|}{GVC}                      & \multicolumn{3}{c|}{FCC}                      & \multicolumn{3}{c|}{ECB+}                     & \multicolumn{3}{c|}{CD2CR}                    \\ \hline
\multicolumn{1}{|l|}{Method}                             & R             & P             & F1            & R             & P             & F1            & R             & P             & F1            & R             & P             & F1            & R             & P             & F1            \\ \hline
\multicolumn{1}{|l|}{\citet{barhom2019revisiting}}       & 81.8          & 77.5          & 79.6          & 81.0 & 66.0          & 72.7          & 17.9          & \textbf{88.3} & 29.8          & 66.8          & 75.5          & 70.9          & -             & -             & -             \\
\multicolumn{1}{|l|}{\citet{barhom2019revisiting}*}      & -             & -             & -             & -             & -             & -             & 36.0          & 83.0          & 50.2          & -             & -             & -             & -             & -             & -             \\
\multicolumn{1}{|l|}{\citet{bugert2020crossdocument}*}   & 71.8          & 81.2          & 76.2          & 49.9          & 73.6          & 59.5          & 38.3          & 70.8          & 49.7          & -             & -             & -             & -             & -             & -             \\
\multicolumn{1}{|l|}{\citet{cattan2020streamlining}}     & 82.1          & 82.7          & 82.4          & -             & -             & -             & -             & -             & -             & 70.7          & 74.8          & 72.7          & 57.0          & 35.0          & 44.0          \\
\multicolumn{1}{|l|}{\citet{yu2020paired}}               & 86.1          & 84.7          & 85.4          & -             & -             & -             & -             & -             & -             & -             & -             & -             & -             & -             & -             \\
\multicolumn{1}{|l|}{\citet{caciularu2021crossdocument}} & 84.9          & \textbf{87.9} & \textbf{86.4} & -             & -             & -             & -             & -             & -             & 82.5          & \textbf{81.7} & 82.1 & -             & -             & -             \\ \hline
\multicolumn{1}{|l|}{Our Approach$^{-}$}     & 84.9          & 82.4          & 83.6          & 67.2          & 81.1          & 73.5          & 47.9          & 68.7          & 56.5          & 84.8          & 76.2          & 80.3          & 67.7          & 72.8          & 70.2          \\
\multicolumn{1}{|l|}{Our Approach$^{+}$}                      & \textbf{85.6} & 87.7          & \textbf{86.6}          & \textbf{82.2}          & \textbf{83.8} & \textbf{83.0} & \textbf{61.6} & 65.4          & \textbf{63.5} & \textbf{85.1} & 80.6          & \textbf{82.8}          & \textbf{77.4} & \textbf{79.7} & \textbf{78.5} \\ \hline
\end{tabular}
\caption{Evaluation Results using $B^3$. For our approaches, ($^{+}$)/($^{-})$ indicates usage of discourse or only a single sentence respectively. Methods marked with * perform all pairwise comparisons without pruning.}
\label{fig:gvc+fcc_eval}
\end{table*}

\section{Related Work}
\paragraph{Cross-Document Coreference}
Many  CDCR algorithms use hand engineered event features to perform classification. Such systems have a low pairwise classification cost and therefore ignore the quadratic scaling and perform no pruning ~\citep{bejan-harabagiu-2010-unsupervised,yang2015hierarchical,vossen2017identity,bugert2020crossdocument}. Other such systems choose to include document clustering to increase precision, which can be done with very little tradeoff for the ECB+ corpus~\citep{lee-etal-2012-joint,cremisini-finlayson-2020-new}.

\citet{kenyon-dean-etal-2018-resolving} explore an approach that avoids pairwise classification entirely, instead relying purely on representation learning and clustering within an embedding space. They propose a novel distance based regularization term for their classifier that encourages representations that can be used for clustering. This approach is more scalable than pairwise classification approaches, but its performance lags behind the state-of-the-art as it cannot use pairwise information.

Most recent systems use neural models for pairwise classification~\citep{barhom2019revisiting, cattan2020streamlining,meged2020paraphrasing, zeng-etal-2020-event, yu2020paired, caciularu2021crossdocument}. These algorithms each use document clustering, a pairwise neural classifier to construct distance matrices within each topic, and agglomerative clustering to compute the final clusters. Innovation has focused on the pairwise classification stage, with variants of document clustering as the only pruning option. \citet{caciularu2021crossdocument} sets the previous state of the art for both events and entities in ECB+ using a cross-document language model with a large context window to cross-encode and classify a pair of mentions with the full context of their documents.

\paragraph{Other Tasks}
\citet{lee-etal-2018-higher} introduces the concept of a ``coarse-to-fine" approach in single document entity coreference resolution. The architecture utilises a bi-linear scoring function to generate a set of \textit{likely} antecedents, which is then passed through a more expensive classifier which performs higher order inference on antecedent chains. Our work extends to multiple documents the idea of using a high recall but low precision pruning function combined with expensive pairwise classification to balance recall, precision, and runtime efficiency.

\citet{wu-etal-2020-scalable} use a similar architecture to ours to create a highly scalable system for zero-shot entity linking. Their method treats entity linking as a ranking problem, using a bi-encoder to retrieve possible entity mentions and then re-ranking the candidate mentions using a cross-encoder. Their results confirm that such architectures can deliver state of the art performance while achieving tremendous scale. However, in coreference resolution, mentions can have one, many, or no coreferring mentions which makes treating it as a ranking problem non-trivial and necessitates the novel training and inference processes we propose.

\begin{figure}[htb]
    \centering
    \removelatexerror
    \begin{algorithm}[H]
\SetAlgoLined
 $M_{e}$: mentions\;
 $s(\cdot, \cdot)$: bi-encoder scorer\;
 $p(\cdot, \cdot)$: cross-encoder scorer\;
 pairs $\gets$ nearestNeighborPairs($M_{e}$, $s(\cdot, \cdot)$)\;
 likelyPairs $\gets$ scoreAndSort(pairs, $p(\cdot, \cdot)$)\;
 C $\gets$ InitializeClustersAsSingletons($M_{e}$)\; 
 \For{pair $\gets$ likelyPairs}{
  $(e_i, e_j) \gets pair$\;
  $c_i \gets$ currentCluster(C, $e_i$)\;
  $c_j \gets$ currentCluster(C, $e_j$)\;
  \eIf{clusterScore($c_i$, $c_j$) $>$ 0.5}{
   C $\gets$ mergeClusters(C, $c_i$, $c_j$)
   }{
   continue\;
  }
 }
 return C\;
 \caption{Inference Algorithm}
\end{algorithm}
    \caption{Clustering algorithm used at inference time}
    \label{fig:merge_algo}
\end{figure}
\section{Model}
Our system is trained in multiple stages and evaluated as a single pipeline. First, we train the encoder for the pruning model to define our latent embedding space. Then, we use this model to sample training data for a pairwise classifier which performs binary classification for coreference. Our complete pipeline retrieves candidate pairs from the attentional state, classifies them using the pairwise classifier, and performs a variant of the agglomerative clustering algorithm proposed by \citet{barhom2019revisiting} to form the final clusters, as laid out in Figure \ref{fig:merge_algo}.
\subsection{Candidate Retrieval}
\paragraph{Encoding Setup}
 We feed the sentences from a window surrounding the mention sentence to a fine-tuned BERT architecture initialized from RoBERTA-large pre-trained weights~\citep{devlin2019bert, liu2019roberta}. A mention is represented as the concatenation of the token-level representations at the boundaries of the mention, following the span boundary representations used by \citet{lee2017endtoend}.
\paragraph{Optimization}
Similar to \citet{kenyon-dean-etal-2018-resolving}, the network is trained to perform a multi-class classification problem where the classes are labels assigned to the gold coreference clusters, which are the connected components of the coreference graph. Rather than adding distance based regularization, we instead optimize the distance metric directly by using the inner product as our scoring function. 

Before each epoch, we construct the representation of each mention $y_{m_i}$ with the encoder from the previous epoch. Each gold coreference cluster $y_{c_i}$ is represented as the centroid of its component mentions $c_i$:
\begin{equation}
    y_{c_i} = \frac{1}{\mid c_i\mid}\sum_{y_{m_i}\in c_i}\!y_{m_i}
\end{equation}
The score $s_o$ of a mention $m_i$ for a cluster $c_i$ is simply the inner product between this cluster representation and the mention representation:
\begin{equation}
s_o(m_i, c_i) = y_{m_i} \cdot y_{c_i}      
\end{equation}
Using this scoring function, the model is trained to predict the correct cluster for a mention with respect to sampled negative clusters. We combine random in-batch negative clusters with hard negatives from the top 10 predicted gold clusters for each training sample in the batch, following \citet{gillick2019learning}. For each mention $m_i$ with true cluster $c'$ and negative clusters $B$, the loss is computed using Categorical Cross Entropy loss on the softmax of our score vector, which we express as:
\begin{equation}
    L(m_i, c') = - s_o(m_i, c') + \log\sum_{c_i \in B} \exp(s_o(m_i, c_i))
\end{equation}
This loss function can be interpreted intuitively as rewarding embeddings which form separable dense mention clusters according to their gold coreference labels. The left term in our loss function acts as an attractive component towards the centroid of the gold cluster, while the right term acts as a repulsive component away from the centroids of incorrect clusters. The repulsive component is especially important for singleton clusters, whose centroids are by definition identical to their mention representations.
\paragraph{Inference}
Unlike previous work using the bi-encoder architecture, our inference task is distinct from our training task. Since our training task requires oracle knowledge of the gold coreference labels, it cannot be performed at inference time. However, since the embedding model is optimized to place all mentions near their centroids, it implicitly places all mentions of the same class close to one another even when that class is unknown. Therefore, the set of K nearest mentions within this space is made up of coreferences and references to highly related entities/events such as shown in Table \ref{fig:examples}, which models an attentional state made up of entities/events explicitly and implicitly in focus~\citep{grosz-sidner-1986-attention}.

Compared to document clustering, this approach can prune aggressively without disregarding any links. The encoding step scales linearly and old embeddings do not need to be recomputed if new documents are added. Importantly, no pairs are disregarded a priori when we compute the nearest neighbor graph and this efficient computation can scale to millions of points using GPU-enabled nearest neighbor libraries like FAISS~\citep{JDH17}, which we use for our implementation.
\subsection{Pairwise Classifier}
\paragraph{Classification Setup}\label{cross-encoder}
For pairwise classification, we use a transformer with cross-attention between pairs.  This follows prior work demonstrating that such encoders  pick up distinctions between classes which previously required custom logic~\citep{yu2020paired}. 
Our use of cross-attention is also motivated by
discourse coherence theory. \citet{grosz-1978-focusing-dialog} highlights that, within an attentional state, the importance to coreference of a mention's features depends heavily on the features of the mention it is being compared to.

The cross encoder is a fine-tuned BERT architecture starting with RoBERTA-large pre-trained weights. For a mention pair $(e_i, e_j)$, we build a pairwise representation by feeding the following sequence to our encoder, where $S_i$ is the sentence in which the mention occurs and $w$ is the maximum number of sentences away from the mention sentence we include as context:
\begin{quote}
    \small
    \centering
    $\langle s \rangle~S_{i-w}$\ldots$S_{i}$\ldots$S_{i+w}~\langle/s\rangle\langle s \rangle~S_{j-w}$\ldots$S_{j}$\ldots$S_{j+w}~\langle/s\rangle$
\end{quote}

Each mention is represented as $v_{e_i}$ which is the concatenation of the representations of its boundary tokens, with the pair of mentions represented as the concatenation of each mention representation and the element-wise multiplication of the two mentions:
\begin{equation}
    v_{(e_i, e_j)} = [v_{e_i} , v_{e_j} , v_{e_i} \odot v_{e_j} ]
\end{equation}
This vector is fed into a multi-layer perceptron and we take the softmax function to get the probability that $e_i$ and $e_j$ are coreferring.
\paragraph{Training Pair Generation}
We use K nearest neighbors in the bi-encoder embedding space to generate training data for the pairwise classifier. This provides the training data a similar distribution of positives and negatives as the classifier will likely see at inference time, but also serves to sample only positive and hard negative pairs.

These negatives are those that the bi-encoder was unable to separate clearly in isolation, which makes them prime candidates for more expensive cross-comparison. At training time, the selection of hyperparameter K is used to balance the volume of training data with the difficulty of negative pairs.
\paragraph{Optimization}
 Once the training data has been generated, we simply train the classifier in a binary setup to classify a pair as either coreferring or non-coreferring. As with prior work, we optimize our pairwise classifier using binary cross-entropy loss. 
\subsection{Clustering}
At inference time, we use a modified form of the agglomerative clustering algorithm designed by \citet{barhom2019revisiting} to compute clusters, as described in Figure \ref{fig:merge_algo}. We do not perform mention detection, so our method relies on gold mentions or a separate mention detection step. First, it generate pairs of mentions using K nearest neighbor retrieval within our embedding space. Each of these pairs is run through the trained cross-encode and all pairs with a probability of less than 0.5 are removed. Pairs are then sorted by their classification probability and clusters are merged greedily. 

Following \citet{barhom2019revisiting}, we compute the score between two clusters as the average score between all mention pairs in each cluster. However, since we only compare two clusters that share a local edge, we do this without computing the full pairwise distance matrix.
\begin{table*}[]
\centering
\small
\begin{tabular}{lc|ccc|ccc|ccc|ccc|}
\cline{3-14}
                               & \multicolumn{1}{l|}{}         & \multicolumn{12}{c|}{Test Dataset}                                                                                           \\ \cline{3-14} 
                               &                               & \multicolumn{3}{c|}{ECB+}   & \multicolumn{3}{c|}{GVC}    & \multicolumn{3}{c|}{FCC}    & \multicolumn{3}{c|}{Harmonic Mean} \\ \hline
\multicolumn{1}{|l|}{Model}    & Train Dataset                 & R    & P    & F1            & R    & P    & F1            & R    & P    & F1            & R       & P      & F1              \\ \hline
\multicolumn{1}{|l|}{Baseline} & ECB+                          & 71.8 & 81.2 & 76.2          & 40.1 & 50.3 & 44.6          & 21.6 & 71.0 & 33.1          & 35.2    & 64.8   & 45.6            \\
\multicolumn{1}{|l|}{Ours}     &                               & 87.1 & 85.3 & \textbf{86.2} & 59.3 & 70.7 & 64.5          & 28.5 & 78.0 & 41.7          & 47.3    & 77.6   & 58.8            \\ \hline
\multicolumn{1}{|l|}{Baseline} & \multirow{2}{*}{FCC}          & 22.1 & 89.0 & 35.4          & 6.4  & 82.9 & 11.9          & 38.3 & 70.8 & 49.7          & 13.2    & 80.2   & 22.6            \\
\multicolumn{1}{|l|}{Ours}     &                               & 88.3 & 19.3 & 31.7          & 63.3 & 29.0 & 39.8          & 51.7 & 73.2 & 60.6          & 64.6    & 30.0   & 41.0            \\ \hline
\multicolumn{1}{|l|}{Baseline} & \multirow{2}{*}{GVC}          & 78.9 & 63.5 & 70.4          & 49.9 & 73.6 & 59.5          & 31.0 & 62.6 & 41.5          & 46.2    & 66.2   & 54.4            \\
\multicolumn{1}{|l|}{Ours}     &                               & 88.4 & 44.2 & 58.9          & 78.6 & 78.8 & 78.7          & 46.1 & 48.5 & 47.3          & 65.6    & 53.6   & 59.0            \\ \hline
\multicolumn{1}{|l|}{Baseline} & \multirow{2}{*}{ECB+ \& FCC}  & 71.8 & 77.2 & 74.4          & 41.2 & 46.5 & 43.7          & 31.0 & 71.6 & 43.3          & 42.6    & 62.0   & 50.5            \\
\multicolumn{1}{|l|}{Ours}     &                               & 83.3 & 86.2 & 84.7          & 59.0 & 70.8 & 64.4          & 49.2 & 87.0 & 62.9          & 60.9    & 80.6   & 69.4            \\ \hline
\multicolumn{1}{|l|}{Baseline} & \multirow{2}{*}{ECB+ \& GVC}  & 78.1 & 68.5 & 73.0          & 46.4 & 40.0 & 43.0          & 39.2 & 50.0 & 43.9          & 50.1    & 50.3   & 50.2            \\
\multicolumn{1}{|l|}{Ours}     &                               & 84.1 & 85.5 & 84.8          & 80.5 & 87.0 & \textbf{83.6} & 26.6 & 78.5 & 39.7          & 48.4    & 83.5   & 61.3            \\ \hline
\multicolumn{1}{|l|}{Baseline} & \multirow{2}{*}{GVC \& FCC}   & 78.2 & 50.6 & 61.4          & 48.8 & 60.7 & 54.1          & 61.0 & 39.6 & 48.0          & 60.4    & 48.8   & 54.0            \\
\multicolumn{1}{|l|}{Ours}     &                               & 94.2 & 19.4 & 32.2          & 82.2 & 75.3 & 78.6          & 54.7 & 77.2 & \textbf{64.0} & 73.1    & 38.6   & 50.5            \\ \hline
\multicolumn{1}{|l|}{Baseline} & \multirow{2}{*}{All Datasets} & 87.2 & 32.3 & 47.1          & 70.7 & 29.6 & 41.7          & 50.8 & 42.6 & 46.3          & 66.2    & 34.0   & 44.9            \\
\multicolumn{1}{|l|}{Ours}     &                               & 83.4 & 84.0 & 83.7          & 70.8 & 86.7 & 78.0          & 49.1 & 72.3 & 58.6          & 64.6    & 80.5   & \textbf{71.6}   \\ \hline
\end{tabular}
\caption{Cross-Evaluation of our approach compared to \citet{bugert2020crossdocument} using the $B^3$ metric}
\label{fig:cross-eval}
\end{table*}
\section{Experiments}
We perform an empirical study across 3 event and 2 entity English cross-document coreference corpora.
\subsection{Datasets}
Here we briefly cover the properties of each corpus we evaluate on. For a more thorough breakdown of corpus properties for event CDCR, see \citet{bugert2020crossdocument}.
\paragraph{Event Coreference Bank Plus (ECB+)}
Historically, the ECB+ corpus has been the primary dataset used for evaluating CDCR. This corpus is based on the original Event Coreference Bank corpus from \cite{bejan-harabagiu-2010-unsupervised}, with entity annotations added in \citet{lee-etal-2012-joint} to allow joint modeling and additional documents added by \citet{cybulska-vossen-2014-using}. By number of documents, it is the largest corpus we evaluate on with 982 articles covering 43 diverse topics. It contains 26,712 coreference links between 6,833 event mentions and 69,050 coreference links between 8289 entity mentions.

\paragraph{Gun Violence Corpus (GVC)}
The Gun Violence Corpus was introduced by \citet{vossen-etal-2018-dont} to present a greater challenge for CDCR by curating a corpus with high similarity between all mentions and documents covered. All 510 articles in the dataset cover incidents of gun violence and are lexically similar which presents a greater challenge for document clustering. It contains 29,398 links between 7,298 event mentions.

\paragraph{Football Coreference Corpus (FCC)}
\citet{Bugert2020BreakingTS} introduced the Football Coreference Corpus in order to evaluate the ability for CDCR systems to identify event coreference across sub-topics. It contains 451 documents covering Football tournaments, where articles covering one tournament often refer to events from other tournaments. While it is the smallest corpus in terms of document size, it has the largest number of coreference links of any dataset we evaluate on with 145,272 coreference links between 3,563 event mentions. \citet{bugert2020crossdocument} re-annotates this corpus at the token level and adds entity labels to enable easier validation between FCC and ECB+.

\paragraph{Cross-Domain Cross-Document Coreference Corpus (CD2CR)}
\citet{ravenscroft2021cd2cr} presents a dataset which evaluates the ability for CDCR models to work across domains which vary significantly in style and vocabulary. It contains 918 documents documents, made up of a 459 pairs of a scientific paper and a newspaper article covering the paper. These articles cover a variety of topics, but since documents come in automatically discovered pairs existing evaluations use the gold document pairs. It contains 13,169 links between 3102 entity mentions.

\subsection{Evaluation and Results}
 All models are implemented in PyTorch~\citep{NEURIPS2019_9015} and optimized with Adam~\citep{Kingma2015AdamAM}. Training the whole pipeline takes one day on a single Tesla V100 GPU. For ECB+, we use the data split used by \citet{Cybulska2015BagOE}. For both FCC and GVC, we use the data splits used by \citet{bugert2020crossdocument}. For CD2CR, we use the splits used by \citet{ravenscroft2021cd2cr}. We compare the $B^3$ metric, since it is reported by baselines for all corpora and has the fewest applicable downsides identified by \citet{moosavi-strube-2016-coreference} since we do not perform mention identification (a full table of metrics for our corpus tailored systems can be found in Appendix \ref{fig:FULL-METRICS}). We use a context window size of 5 sentences during candidate retrieval and of 3 sentences during pairwise classification for all experiments. For corpus tailored evaluations, we retrieve 15 pairs for each mention at training time and 5 pairs at inference time. For cross corpus evaluations, we retrieve 5 pairs for each mention for both training and inference.
\paragraph{ECB+}
 Our approach achieves a new state of the art result on ECB+, which is the most widely used CDCR dataset. Our results improve on \citet{caciularu2021crossdocument} by 0.2 F1 points for events and 0.7 F1 points for entities. This result is particularly noteworthy since document clustering can be performed nearly perfectly for the ECB+ dataset~\citep{barhom2019revisiting} and there are no inter-cluster links~\citep{bugert2020crossdocument}. 

Given that document clustering has almost no downside for ECB+ and \citet{caciularu2021crossdocument} uses a cross-encoder architecture with a much wider context window for classification, we largely credit the increased performance on ECB+ dataset to the benefits of hard sampling using our attentional state neighborhoods.
\paragraph{GVC \& FCC}
We evaluate the broader applicability of our model for event CDCR by applying it to the FCC and GVC datasets. Each aim to address elements of real world event CDCR overlooked by ECB+. These datasets only annotate events, preventing joint modeling of events and entities. This negatively impacts \citet{barhom2019revisiting} which was designed as a joint method, but requires no changes to our architecture.

Our approach improves over the state of the art by 11.3 F1 points for the GVC dataset and by 13.1 F1 points for the FCC dataset. It is worth noting that the previous state-of-the-art was split between these datasets, with document clustering benefiting GVC and harming FCC performance. Our approach improves on the results for both datasets without modification, unifying the state-of-the-art under one approach.
\paragraph{CD2CR}
CD2CR presents a unique challenge with coreference links which span two domains with very different linguistic properties: academic text and science journalism. While one might expect that this linguistic diversity could cause our pruning method to struggle to retrieve pairs across domains, our method proves robust to this challenge with a 34.5 F1 point improvement over the state-of-the-art. This is especially significant as CD2CR previously used a highly corpus-tailored document linking algorithm that relied on data such as DOI matching and author name and affliation matching since document clustering algorithms used for ECB+ are a bad fit for CD2CR due to the within-topic lexical diversity. This highlights how flexible our method is compared to document clustering.

\paragraph{Event Cross-Dataset Evaluation}
We evaluate the robustness of our learned models by training and evaluating across the multiple event datasets. \citet{bugert2020crossdocument} propose cross-corpus training as a treatment to produce more generally effective models, since downstream corpora are unlikely to match any specific CDCR corpus. We follow their cross-corpus evaluation and present the results for this cross-evaluation in Table \ref{fig:cross-eval}.

For models trained on the train split from a single corpus, we see significant performance loss when evaluated on test splits from other corpora as is expected. However, we see vastly improved generalizability with our approach when trained on a single corpus compared to the baseline set by \citet{bugert2020crossdocument}.

To evaluate the ability of our model to learn from multiple corpora at once, we train our pipeline on combinations of multiple datasets. Datasets are combined naively by using all documents and mentions from the train split of each corpus.

Interestingly, our performance improves on FCC and GVC when training our model with two out of three datasets for both GVC and FCC. We achieve our best results on FCC when GVC training data is added and our best results on GVC when ECB+ data is added. This signals that there is potential for further improvement of the model trained on all datasets by exploring what causes the performance decrease with the introduction of the third dataset in these two cases.

Most importantly, our model trained across all datasets shoes improved generalizability across each dataset, sacrificing 2.9, 5.0, and 4.9 F1 points compared to our state-of-the-art corpus tailored models for ECB+, GVC, and FCC respectively. This is a 4.27 point F1 decrease on average compared to 16.7 F1 points for the baseline, suggesting that our model more effectively adapts to the varying feature importance across corpora shown by \citet{bugert2020crossdocument}. For use in downstream systems, this model variant makes it feasible variety of downstream corpora without fine-tuning, which is especially important since the majority of downstream tasks lack coreference annotations for fine-tuning.
\begin{table}[]
\centering
\begin{tabular}{|l|ccc|}
\hline
 Pairwise Classifier & R & P & F1\\ \hline
 \citet{barhom2019revisiting} & 76.2 & 70.7 & 73.4 \\
 \citet{yu2020paired} & 84.4 & 81.4 & 82.9 \\ 
Discourse Cross-Encoder & 87.1 & 85.3 & 86.2 \\ 
Oracle Model & 96.3 & 1.0 & 98.1 \\\hline   

\end{tabular}
\caption{Candidate Retrieval with Alternate Classifiers evaluated on ECB+ using $B^3$}
\label{fig:substitute_study}
\end{table}
\begin{table*}[]
\centering
\begin{tabular}{l|ccc|ccc|}
\cline{2-7}
                                          & \multicolumn{3}{c|}{Events} & \multicolumn{3}{c|}{Entities} \\ \hline
\multicolumn{1}{|l|}{Model Variant}           & R       & P       & F1      & R        & P        & F1      \\ \hline
\multicolumn{1}{|l|}{Our Approach with Discourse}        & 87.1    & 85.3    & 86.2    & 84.1     & 77.6     & 80.7    \\
\multicolumn{1}{|l|}{\hspace{5pt}$-$ Time and Location} & 84.5    & 85.9    & 85.2    & 82.6     & 79.0     & 80.7    \\
\multicolumn{1}{|l|}{\hspace{5pt}$-$ Coreference}       & 85.2    & 86.0    & 85.6    & 83.5     & 72.9     & 77.8    \\
\multicolumn{1}{|l|}{\hspace{5pt}$-$ All Entities}      & 82.0    & 87.9    & 84.9    & 81.4     & 73.2     & 77.1    \\
\multicolumn{1}{|l|}{\hspace{5pt}$-$ All Events}        & 88.2    & 82.3    & 85.1    & 81.4     & 80.5     & 81.0    \\
\multicolumn{1}{|l|}{Our Approach without Discourse}     & 84.4    & 81.4    & 82.9    & 84.1     & 69.4     & 76.0    \\ \hline
\end{tabular}
\caption{Masking Study of Discourse Cross-Encoder. Masking is applied only to sentences from the context window, leaving the sentence where the mention occurs fully unmasked. ($^{+}$)/($^{-})$ indicates usage of discourse or only a single sentence respectively.}
\label{fig:ablation}
\end{table*}

\section{Analysis}
We analyze the components of our model in isolation to explain the sources of our significant performance gains and bottlenecks which still exist.
\subsection{Candidate Retrieval Isolation}\label{biencoder_isolation}
We evaluate our pruning method with alternate classifiers in Table \ref{fig:substitute_study}. For these experiments, we fetch 5 nearest neighbor pairs for each mention.

We define the upper bound performance of our pruning method by performing an oracle study where the pruned pairs are passed pairwise classifier that has access to gold labels. Despite using only 5 nearest neighbors the system achieves a recall of 96.3, resulting in an upper-bound F1 of 98.1. Future works can use our pruning method with improved pairwise classification methods without concern since the pruning method delivers near perfect results with an oracle pairwise classifier.

We isolate the benefits of our pairwise classification approach by using our pruning model with the pairwise classifiers of \citet{barhom2019revisiting} and the trigger-only variant of \citet{yu2020paired}. The resulting performance is worse than that of our work, indicating that the pairwise classification model we utilize also plays an important role in our results. Our approach varies from \citet{yu2020paired} by using a hard negative training approach and local discourse features, leading us to believe these are the primary beneficial factors.

\subsection{Discourse Context Ablation Study}\label{ablation}
Both our work and the prior state-of-the-art~\citep{caciularu2021crossdocument} utilize discourse features during pairwise comparison, which significantly improves performance compared to just a single sentence of context. However, it is not well understood what features of local discourse are valuable to CDCR. We analyze the contributions of local discourse information through two ablation studies.

We first evaluate the sensitivity of our model to hyperparameter $w$, the number of sentences surrounding each mention included as context, by keeping a fixed bi-encoder and training 4 separate cross-encoders from $w=0$ up until $w=3$. Due to our model's 512 token limit, we do not evaluate over $w=3$. The results of this ablation, shown in Table \ref{fig:ablation_w}, demonstrate that each increase in window size increases performance, with diminishing returns.

To understand which local discourse features contribute to this improvement, we study three special types of token from the surrounding discourse: times, locations, and coreferences. Time and location within a sentence has been used in past work using semantic role labeling~\citep{barhom2019revisiting, bugert2020crossdocument} and coreferring tokens are intuitively informative as they provide additional information about the same event/entity. By including local discourse, 21\%, 11\%, 29\% of events and 18\%, 9\%, 34\%  entities gain access to new time, location, and coreference information respectively. For example, consider the following text: 
\begin{quote}
    A strong earthquake struck \textit{Indonesia's Aceh province} on \textit{Tuesday}. Many houses were \textbf{damaged} and dozens of villagers were injured.
\end{quote}

While the event "damaged" is ambiguous with only the context of a single sentence, it becomes much more specific when contextualized with the previous sentence which contains both a time and a location for the event. We evaluate our system with tokens of these types masked from the local discourse with results reported in Table \ref{fig:ablation}.

For events, both masking time and location (-1.0 F1) and masking coreference (-0.6 F1) in the local discourse significantly harms performance . However, only within-document coreference seems to majorly impact entity resolution (-2.9 F1). Both events and entities are more impacted by masking all entities (-1.3 F1 for events, -3.6 for entities) than they are by masking all events (-1.1 F1 for events, +0.3 F1), which matches the expectation that the greater degree of polysemy for event tokens makes them less discriminative.

\begin{table}[]
\centering
\begin{tabular}{|l|lll|}
\hline
 $w$ & R & P & F1\\ \hline
 0 & 84.4 & 81.4 & 82.9 \\
 1 & 83.4 & 86.5 & 84.9 \\ 
 2 & 83.1 & 87.7 & 85.4 \\ 
 3 & 87.1 & 85.3 & 86.2 \\\hline   

\end{tabular}
\caption{Ablation on cross-encoder context window $w$ evaluated on ECB+ using $B^3$}
\label{fig:ablation_w}
\end{table}

\section{Conclusion and Future Work}
In this work, we presented a two-step method for resolving cross-document event and entity coreference inspired by discourse coherence theory. We achieved state-of-the-art results on 3 event and 2 entity CDCR datasets, unifying the previously fractured CDCR space with a single model. We further improve applicability by training across corpora, presenting a model which can be used for downstream tasks that lack coreference annotations for fine-tuning. We demonstrated that our pruning method offers high upper bound performance and that both stages of our model contribute to our state-of-the-art results. Finally, we explained contributions of local discourse features when cross-encoding for coreference resolution.

We identify 3 areas of future work: 
\begin{itemize}
  \item Using knowledge distillation to further improve scalability. \citet{wu-etal-2020-scalable} demonstrate that much of the quality gain from cross-encoding can be transferred to a bi-encoder through knowledge distillation, which could have the potential to remove pairwise classification altogether. 
  \item Pairing alternate models for pairwise classification with the bi-encoder candidate pair generator. Our candidate pair generator is unlikely to become a recall bottleneck, so future efforts in CDCR should focus primarily on improving the accuracy of pairwise classification.
  \item Integrating CDCR into a wider range of tasks. Our work is robust to a wide variety of data, but it is still unknown which cross-document tasks benefit the most from coreference information.
\end{itemize}

\paragraph{Acknowledgments} We are very grateful to Marissa Mayer and Enrique Muñoz Torres of Sunshine Products for their steadfast encouragement and generous financial support of this project,
to Ken Scott and Annie Luu of Sunshine Products for logistics assistance, and to the anonymous reviewers and the Stanford NLP group for their helpful feedback.
\bibliography{emnlp2020}
\bibliographystyle{acl_natbib}
\newpage
\pagebreak[4]
\clearpage

\appendix

\section{Full Metrics Report}\label{fig:FULL-METRICS}
In Table \ref{fig:FULL-METRICS-FIG}, we present a table of the commonly used metrics for evaluating CDCR systems for each of our corpus-tailored systems for the sake of future comparisons.
\begin{table*}[]
\centering
\small
\begin{tabular}{cc|ccc|ccc|ccc|c|ccc|}
\cline{3-15}
\multicolumn{1}{l}{}                             & \multicolumn{1}{l|}{} & \multicolumn{13}{c|}{Metric}                                                                                                                              \\ \cline{3-15} 
\multicolumn{1}{l}{}                             &                       & \multicolumn{3}{c|}{MUC} & \multicolumn{3}{c|}{$B^3$} & \multicolumn{3}{c|}{CEAFe} & \multicolumn{1}{l|}{CoNLL} & \multicolumn{3}{c|}{LEA} \\ \hline
\multicolumn{1}{|c|}{Type}                       & Dataset               & R      & P      & F1     & R            & P            & F1          & R       & P       & F1     & F1                         & R      & P      & F1     \\ \hline
\multicolumn{1}{|c|}{\multirow{3}{*}{Event}}    & ECB+                  & 87.0   & 88.1   & 87.5   & 85.6         & 87.7         & 86.6        & 80.3    & 85.8    & 82.9   & 85.7                       & 74.9   & 73.2   & 74.0   \\ 
\multicolumn{1}{|c|}{}                           & GVC                   & 91.8   & 91.2   & 91.5   & 82.2         & 83.8         & 83.0        & 75.5    & 77.9    & 76.7   & 83.7                       & 79.0   & 82.3   & 80.6   \\ 
\multicolumn{1}{|c|}{}                           & FCC                   & 86.4   & 75.7   & 80.7   & 61.6         & 65.4         & 63.5        & 39.1    & 65.3    & 48.9   & 64.4                       & 47.2   & 57.0   & 51.6   \\ \hline
\multicolumn{1}{|c|}{\multirow{2}{*}{Entity}} & ECB+                  & 88.2   & 89.5   & 88.9   & 85.1         & 80.6         & 82.8        & 75.7    & 73.1    & 74.4   & 82.0                       & 77.1   & 74.0   & 75.5   \\  
\multicolumn{1}{|c|}{}                           & CD2CR                 & 78.5   & 96.7   & 86.7   & 77.4         & 79.7         & 78.5        & 43.0    & 69.7    & 53.2   & 72.8                       & 65.0   & 78.8   & 71.2   \\ \hline
\end{tabular}
\caption{MUC, $B^3$, CEAFe, CoNLL, and LEA metrics for each corpus-tailored system}
\label{fig:FULL-METRICS-FIG}
\end{table*}

\end{document}